\documentclass[11pt,a4paper]{article}
\usepackage[hyperref]{acl2019}
\usepackage{times}
\usepackage{latexsym}
\usepackage{graphicx}
\usepackage[utf8]{inputenc}
\usepackage{xcolor}
\usepackage{booktabs}
\usepackage{mathtools}
\usepackage{inconsolata}
\usepackage{enumitem}
\usepackage{amsmath}
\usepackage{amssymb,amsmath,amsthm,enumitem}
\usepackage{bm}
\usepackage{microtype}
\usepackage{dirtytalk}
\usepackage{tikz}
\def\checkmark{\tikz\fill[scale=0.4](0,.35) -- (.25,0) -- (1,.7) -- (.25,.15) -- cycle;} 
\usepackage{float}
\restylefloat{table}
\usepackage{url}

\aclfinalcopy 
\def\aclpaperid{994} 

\title{LSOIE: A Large-Scale Dataset for \\Supervised Open Information Extraction}

\author{Jacob Solawetz~~~~\\
  Roboflow, Inc. \\
  Minneapolis, MN, USA \\
  \texttt{jacob@roboflow.ai} \\\And
  Stefan Larson \\
  Rosegold AI \\
  Ann Arbor, MI, USA \\
  \texttt{stefan@rosegold.ai} \\}

\date{}

\begin{document}
\maketitle
\begin{abstract}
  Open Information Extraction (OIE) systems seek to compress the factual propositions of a sentence into a series of \textit{n}-ary tuples. These tuples are useful for downstream tasks in natural language processing like knowledge base creation, textual entailment, and natural language understanding. However, current OIE datasets are limited in both size and diversity. We introduce a new dataset by converting the \textsc{qa-srl} 2.0 dataset to a large-scale OIE dataset (\textsc{lsoie}). Our \textsc{lsoie} dataset is 20 times larger than the next largest human-annotated OIE dataset. We construct and evaluate several benchmark OIE models on \textsc{lsoie}, providing baselines for future improvements on the task. Our \textsc{lsoie} data, models, and code are made publicly available.\footnote{Our \textsc{lsoie} dataset, models, and code can be found at \url{https://github.com/Jacobsolawetz/large-scale-oie}.}
\end{abstract}

\section{Introduction}

Open Information Extraction (OIE) \cite{Banko:2007:OIE:1625275.1625705} aims to automatically extract all factual propositions of a sentence into a series of $n$-ary tuples. For example, the sentence ``the cook baked and ate the cake'' would produce two extractions representing the two basic propositions of the sentence: (the cook, \textbf{ate}, the cake) and (the cook, \textbf{baked}, the cake). 
In OIE, extraction arguments are required to be contiguous spans from the sentence and the resulting tuple should be intelligible as natural text when read in order. The schema-free nature of OIE provides a flexible framework in which to capture semantic relations between entities in natural language text. Open Information Extraction tuples are useful to a variety of downstream tasks including knowledge base creation \cite{zhang-etal-2019-openki}, textual entailment \cite{levy-etal-2014-focused}, and other natural language understanding tasks \cite{Mausam2016OpenIE}. 

Open Information Extraction relations may be explicitly stated by verbal predicates, or implicitly stated through nominalizations. In this paper, we focus only on explicit extractions.
With the original goal of OIE as web scale information extraction \cite{Banko:2007:OIE:1625275.1625705}, an OIE system can focus solely on explicit extractions because the redundancy of language will inevitably display implicit information elsewhere.

\begin{table}[t]
\small
\centering
\begin{tabular}{llrr}
\toprule
    & \textbf{Domains}
    & \textbf{\#Sent.}
    & \textbf{\#Ext.}
    \\
\midrule
\textsc{oie2016}
      & Wiki, Newswire
      & 3,180
      & 8,477
      \\
\textsc{aw-oie} & Wiki, Wikinews
        & 3,300
        & 17,165 \\
\textsc{lsoie}-wiki
      & Wiki, Wikinews
      & 24,296
      & 56,662
      \\
\textsc{lsoie}-sci
      & Science
      & 47,998
      & 97,550
      \\
\end{tabular}

\begin{tabular}{lrrr}
\toprule
    & \textbf{Ext. / Sent.}
    & \textbf{Vocab}
    & \textbf{Ordered}
    \\
\midrule
\textsc{oie2016}
      & 2.7
      & 13,863
      & 
      \\
\textsc{aw-oie} & 5.2 & 15,853 & \\
\textsc{lsoie}-wiki
      & 2.3
      & 46,617
      & \checkmark
      \\
\textsc{lsoie}-sci
      & 2.0
      & 51,668
      & \checkmark
      \\
\bottomrule

\end{tabular}

\caption{OIE dataset metrics.
Our new dataset \textsc{lsoie} has substantially more text available
than prior work,
and includes a new science domain.
Our dataset conversion process leverages the scope of the \textsc{qa-srl} 2.0 bank and improves upon previous methodology. }
\label{tab:dataset-metrics}
\end{table}

The interest in OIE has grown: both in terms of the types of models that can be applied to tackle OIE \cite{cui-etal-2018-neural, stanovsky-etal-2018-supervised, jiang-etal-2019-improving}, and in terms of the downstream applications to which OIE can be applied \cite{Mausam2016OpenIE, zhang-etal-2019-openki}.
As the interest in OIE grows, however, so too should the scale of the corpora available for training and evaluating OIE models.

\begin{figure*}
  \centering\includegraphics[width=.95\linewidth,height=0.5in]{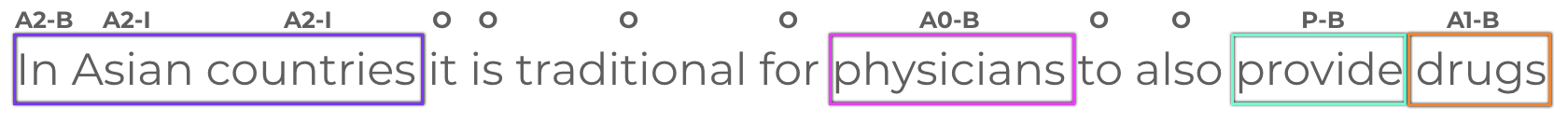}\par
  \caption{
  An example annotated sentence from \textsc{qa-srl} 2.0 \cite{fitzgerald2018qasrl}. In this case, the annotations are derived from the question and answers: 
  - Where does someone {\bf provide} something? {\bf In Asian countries}. Who {\bf provides} something? {\bf physicians}. What is being {\bf provided}? {\bf drugs}.
  The extracted tuple in our new \textsc{lsoie} dataset is (physicians, \textbf{provide}, drugs, in Asian countries).
  \label{fig:example} }
\end{figure*}

In this paper,
we expand the reach and quality of OIE data by developing a new dataset, \textsc{lsoie}, which is built by converting the \textsc{qa-srl bank} 2.0 dataset \cite{fitzgerald2018qasrl} to the task of OIE.
Our new dataset contains almost ten times as many extractions and about 20 times as many sentences as previous OIE datasets built from human annotations (see Table~\ref{tab:dataset-metrics}).
We benchmark \textsc{lsoie} with several models, providing baseline results for future research. Our \textsc{lsoie} dataset, models, and code are publicly available.

\section{Background}
\subsection{OIE Datasets}
Available OIE corpora fall into three categories:
(1) converted from crowdsourcing,
(2) model-derived, and
(3) directly crowdsourced.

\textbf{Converted from crowdsourcing:}
\citet{stanovsky-dagan-2016-creating} created the \textsc{oie2016} dataset by converting the crowd-annotated \textsc{qa-srl} \cite{He2015QuestionAnswerDS} dataset's question-answer pairs to OIE extraction relations. Similarly, \citet{stanovsky-etal-2018-supervised} generated the \textsc{aw-oie} dataset by converting the crowd-annotated Question Answer Meaning Representation (\textsc{qamr}) dataset's question-answer pairs. 

The \textsc{oie2016} and \textsc{aw-oie} datasets were the first datasets used for supervised OIE. These datasets provided the basis for supervised approaches in NLP, but they are small and extractions lack accuracy, as they are converted in the order that question answer pairs appear in the base dataset.

\textbf{Model-derived:}
\citet{cui-etal-2018-neural} and \citet{DBLP:journals/corr/abs-1809-09408} generate large derivative training datasets by running rules-based models and keeping high confidence extractions for downstream tasks.
Similarly, \citet{gashteovski2019opiec} introduce the largest OIE dataset to date (over 340M triples) by deriving extractions from \texttt{MinIE} \citet{gashteovski-etal-2017-minie} with the goal of automatically constructing a knowledge base. 
While model-derived datasets are useful for knowledge base construction, using them for downstream tasks teaches the new model to replicate the behavior of the original, often noisy, base model. 

\textbf{Directly crowdsourced:}
\citet{bhardwaj-etal-2019-carb} point out that the evaluation framework used in \citet{stanovsky-dagan-2016-creating} is rather noisy and the tuple matching algorithm is overly lenient because it only looks at lexical overlap for the whole extraction, ignoring the ordering of arguments.
\citet{bhardwaj-etal-2019-carb} provide an alternative evaluation set that has been crowdsourced specifically for OIE, annotating 1,282 sentences. 
While this dataset is useful for the evaluation of OIE systems, its format differs from other work in OIE - the predicate entry in \textsc{CaRB} \cite{bhardwaj-etal-2019-carb} tuples contains context that is often broken into separate tuples by other OIE systems.

\subsection{The QA-SRL Bank 2.0}

In \textsc{qa-srl}, each predicate-argument relationship in a sentence is labeled manually with a question-answer pair. \citet{fitzgerald2018qasrl} design a large-scale crowdsourcing annotation pipeline to incentivize extensive and accurate coverage. Relative to the original \textsc{qa-srl} annotations \cite{He2015QuestionAnswerDS}, which were collected from 10 hired freelance workers, the new \textsc{qa-srl} dataset achieves similar precision (95.7\% versus 97.5\%) and lower recall (72.4\% versus 86.6\%). Relative to Propbank \cite{10.1162/0891201053630264}, an expert annotation system designed to capture all semantic roles in a sentence. the \textsc{qa-srl} 2.0 authors find that their work 95\% precision and 85\% recall. \citet{fitzgerald2018qasrl} then build a supervised \textsc{qa-srl} parser and extend the reach of their dataset by over-generating new candidate question-answer pairs and passing them through their validation process.

The \textsc{qa-srl} paradigm is well-suited to be a precursor to OIE extractions, as it captures predicate-argument relations in a schema-free way.

\section{The LSOIE Dataset}

Our work expands upon and addresses the shortcomings present in \citet{stanovsky-dagan-2016-creating} and \citet{stanovsky-etal-2018-supervised}.
We apply a similar conversion processes used for \textsc{oie2016} on the \textsc{qa-srl bank} 2.0 dataset.
In addition, we implement novel conversion heuristics to ensure data quality and order arguments. The result is \textsc{lsoie}, an OIE dataset that is much larger and diverse than prior work.

\subsection{LSOIE Conversion Process}

We produce \textsc{lsoie} via conversion from \textsc{qa-srl} in the same manner as \citet{stanovsky-dagan-2016-creating}, with several important changes to adapt their method to the \textsc{qa-srl bank} 2.0.

A \textsc{qa-srl} annotation for a predicate $p$ consists of
a list of questions $Q = \{q_0, \dots, q_n\}$,
and a set of answer spans $A_i = \{a_{i0}, \dots, a_{in_i}\}$
 for each question $q_i$.
For each tuple $(a_0, \dots, a_k) $ in the Cartesian product $ \bigtimes_i^n A_i $,
we produce the extraction tuple $(a_0, p, a_1, \dots, a_k)$.

In our example extraction in Figure~\ref{fig:example}, the target predicate $p$ is \emph{provide}. The list of questions $Q$ is [\emph{Where does someone provide something?}, \emph{Who provides something?}, \emph{What is being provided?}] The list of arguments $A$ is [\emph{In Asian countries}, \emph{physicians}, \emph{drugs}]. The converted extraction tuple is (\emph{physicians}, \emph{provide}, \emph{drugs}, \emph{in Asian countries}).

To ensure data quality and as a result of differences between the original \textsc{qa-srl} dataset
and the \textsc{qa-srl bank} 2.0,
we had to make two important changes to the algorithm:

\makeatletter
\setlength{\@fptop}{0pt}
\makeatother
\begin{figure}[t!]
\hspace*{-.35cm}
  \centering
  \includegraphics[keepaspectratio, width=0.5\textwidth]{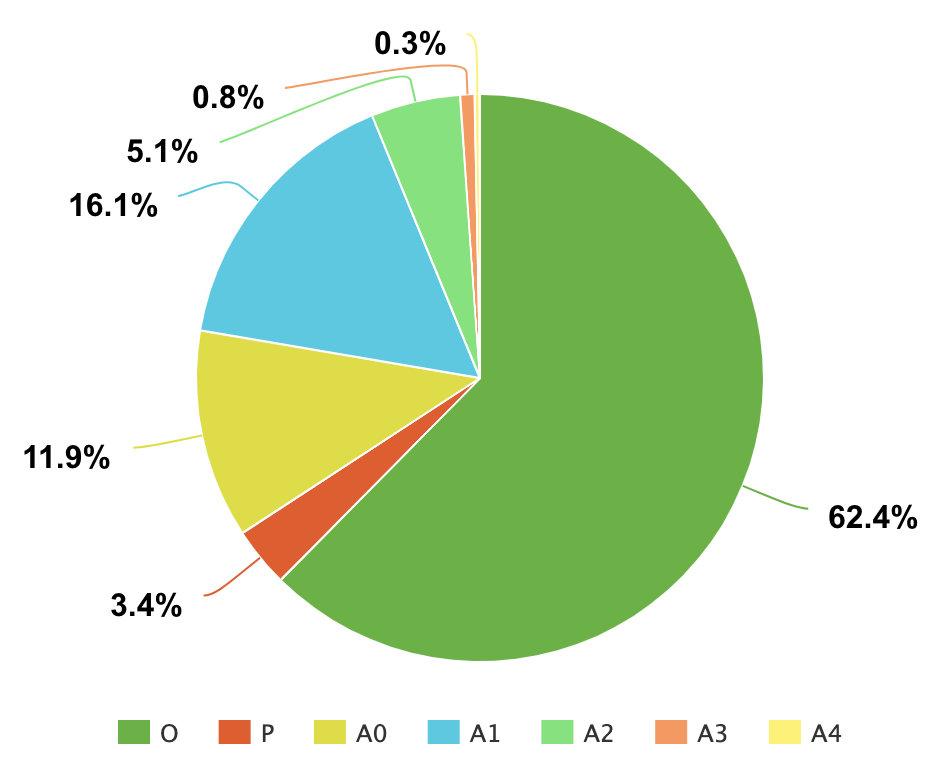}
  \caption{The distribution of token level tags (listed clockwise) in the \textsc{lsoie} dataset. $P$ denotes the extraction's predicate, $A_0$-$A_N$ denote the extraction's arguments, and O denotes that a given token does not belong to the extraction.}
  \label{fig:tag-dist}
\end{figure}

\textbf{Answer Filtering:~}
The original \textsc{qa-srl} dataset has a single set of mutually-exclusive answer spans for each question, written by a single annotator.
In contrast, the \textsc{qa-srl bank} 2.0 has answer judgments from three annotators for each question, some providing answer sets and others marking the questions as invalid.
To consolidate these, we only include questions marked as valid by all three annotators.
Then, for each question, we iteratively draw the longest remaining answer span that does not
overlap with a previously drawn answer span, until there are none left.

In answer filtering, our primary motivation was to clean the raw version of crowd workers' answer responses in the \textsc{qa-srl} 2.0 dataset, where questions can be posed that are not valid or the answer to them is ambiguous. We found it advantageous for dataset quality to require a strict agreement between all annotators. In choosing the longest answer span, we were motivated to not miss relevant portions of the argument, as individual crowd workers occasionally annotated a limited portion of the answer span that did not encapsulate the whole semantic meaning of the derived argument.

\begin{table}[]
    \centering
    \scalebox{0.75}{
    \begin{tabular}{c}
    \toprule
       \emph{Bats are the only mammals that can truly \textbf{fly}.} \\
       (Bats, \textbf{fly}) \\
       \midrule
       \emph{Greece moved up three to be \textbf{ranked} tenth.}\\
       (Greece, \textbf{ranked}, tenth)\\
       \midrule
       \emph{A popular student, in 1915 Mao was}\\ \emph{\textbf{elected} secretary of the Students Society.} \\
       (Mao, \textbf{elected}, secretary of the Students Society, in 1915)\\
       \midrule
       \emph{The proposed amendment already \textbf{passed} both houses in 2011.}\\
       (The proposed amendment, \textbf{passed}, both houses, in 2011)\\
       \midrule
       \emph{In polygynous species, males try to}\\ \emph{\textbf{monopolize} and mate with multiple females.}\\
       (males, \textbf{monopolize}, multiple females)\\
       \midrule
       \emph{Animals \textbf{adapted} to live in the desert are called xerocoles.}\\
       (Animals, \textbf{adapted}, to live in the desert)\\
    \bottomrule
    \end{tabular}}
    \caption{Example sentences with example extractions. Note that only one example extraction is shown here, though a sentence can yield multiple extractions.
    }
    \label{tab:example-data}
\end{table}

\textbf{Argument ~Ordering:~~}
\citet{stanovsky-dagan-2016-creating}'s original algorithm relies on the original, annotator-written order of \textsc{qa-srl} questions, which may or may not produce a sensible argument ordering. Furthermore, in the \textsc{qa-srl bank} 2.0, the original order in which the questions were written is unavailable.

So, to determine argument order, we use a heuristic based on the relative order between answer spans for each question in their source text.
We consider the abstract form of questions, which includes verb tense without information about its lemma.
For a given question $q_i$ in an extraction, let $q_{i_x}$ represent the percentage of predicates in the \textsc{qa-srl bank} 2.0 where the answer span to the generalized version of $q_i$ appears in the $x^{th}$ place relative to other answer spans, according to the natural order of the sentence. For each argument slot in the derived extraction, the answer to the question with the highest probability $q_{i_x}$ of naturally occurring in that slot is chosen as the argument.

In our example extraction in Figure~\ref{fig:example}, the question \emph{Who} \texttt{[predicate]} \emph{something?} precedes \emph{What is being} \texttt{[predicate]}\emph{?} which precedes \emph{Where does someone} \texttt{[predicate]} \emph{something?}, enabling our algorithm to accurately extract argument ordering, which is not available from the natural ordering of the sentence or the ordering of crowd annotations in \citet{fitzgerald2018qasrl}.

\subsection{Dataset Statistics}

We run our updated dataset conversion process over the
directly crowdsourced portion of the
train, development, and test partitions of the \textsc{qa-srl bank} 2.0.
Stratifying the resulting data by domain,
we present the new \textsc{lsoie} corpus in two sections, \textsc{lsoie}-wiki and \textsc{lsoie}-sci.
Dataset statistics are shown in Table~\ref{tab:dataset-metrics}. Example extractions are shown in Table~\ref{tab:example-data}. We provide the distribution of argument, predicate, and null tag labels in Figure~\ref{fig:tag-dist}.
The \textsc{lsoie} corpus expands the scope of \textsc{oie2016} and \textsc{aw-oie} in size, textual diversity, and domain.

\makeatletter
\setlength{\@fptop}{0pt}
\makeatother
\begin{figure}[t!]
\hspace*{-.35cm}
  \centering
  \includegraphics[keepaspectratio, width=0.5\textwidth]{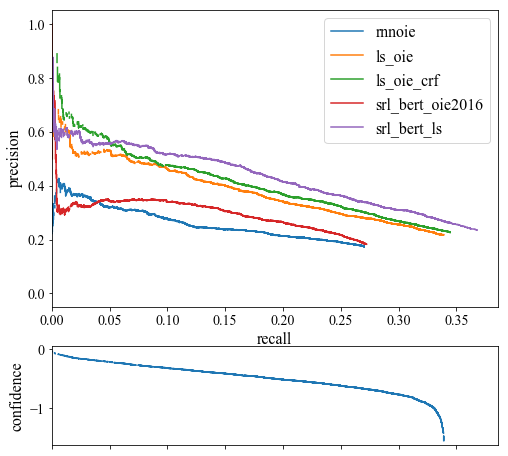}
  \caption{
  Top: performance of Supervised OIE systems on the \textsc{lsoie}-wiki test set.
  Bottom: \texttt{ls\_oie} estimated confidence at each extraction. }
  \label{fig:results-graph}
\end{figure}

\section{Benchmark Evaluation}
\paragraph{Models:}
We evaluate several models on our new \textsc{lsoie} dataset.
Following \citet{stanovsky-etal-2018-supervised},
we model OIE as a supervised learning problem
and format it as BIO tagging with tunable thresholding on extractions.
We benchmark several model variants:

\begin{itemize}[topsep=0pt,partopsep=0pt,itemsep=0pt,parsep=0pt]
    \item \textbf{\texttt{rnnoie}} is a replication of the model in \citet{stanovsky-etal-2018-supervised},
based on a bidirectional LSTM transducer
over GloVe embeddings \citep{Pennington14glove:global}
and learned part-of-speech embedding features.
    \item \textbf{\texttt{ls\_oie}} is a replication of \texttt{rnnoie} trained on \textsc{lsoie}.
    \item \textbf{\texttt{ls\_oie\_crf}} is the same as \texttt{ls\_oie},
but trained end-to-end with a Conditional Random Field on top
to capture BIO transition constraints
and trained to maximize the likelihood of the gold BIO sequence.
    \item \textbf{\texttt{srl\_bert\_ls}}
is based on \texttt{ls\_oie}, but uses BERT \citep{devlin2019bert} as the bidirectional encoder
and the Sentence A / Sentence B embedding feature as the predicate indicator,
inspired by \citet{shi2019simple}.
\item \textbf{\texttt{srl\_bert\_oie2016}} is the same architecture as \texttt{srl\_bert\_ls} but applied to the \textsc{oie2016} data.
    \item \textbf{\texttt{*\_sci}} models were trained with the same architectures applied only to the \textsc{lsoie}-sci training set.
\end{itemize}

\begin{table}[t]
\centering
\scalebox{0.9}{
\begin{tabular}{l|rr|rr}
\toprule
  & \multicolumn{2}{c|}{\textbf{\textsc{lsoie}-wiki}}
  & \multicolumn{2}{c}{\textbf{\textsc{lsoie}-sci}}
  \\
\midrule
  \textbf{Model} & $\boldsymbol{F_1}$ & \textbf{AUC}
   & $\boldsymbol{F_1}$ & \textbf{AUC}
  \\
\midrule
  \texttt{rnnoie}          & .22    &.07&   .26 & .10\\
  \texttt{ls\_oie}           & .28  & .13   & .33 & .18\\
  \texttt{ls\_oie\_crf}      & .29 & .14& .33 & .19\\
  \texttt{srl\_bert\_oie2016}   & .23 & .08 &  .29& .13\\
  \texttt{srl\_bert\_ls}     &  \textbf{.31}& \textbf{.16} & .37 & .21\\
  \texttt{ls\_oie\_sci}      & - &  -  & .34 & .19\\
  \texttt{ls\_oie\_crf\_sci} & -  &  - & .35 & .20\\
  \texttt{srl\_bert\_ls\_sci}    & - & -& \textbf{.38} &  \textbf{.22}
  \\
 \bottomrule
\end{tabular}}
\caption{Modeling results on the \textsc{lsoie} test sets.}
\label{tab:results-table}
\end{table}

\paragraph{Experiments and Evaluation:}
We use
the AllenNLP framework \citep{DBLP:journals/corr/abs-1803-07640}
built on PyTorch \citep{paszke2019pytorch}
to implement, train, and test our models.
We train \texttt{rnnoie} and \texttt{srl\_bert\_oie2016} on \textsc{oie2016} and \texttt{ls\_oie} and \texttt{srl\_bert\_ls} on \textsc{lsoie}-wiki. We also focus the series of models by only training on \textsc{lsoie}-sci. We do not evaluate \texttt{*\_sci} models on \textsc{lsoie}-wiki. We limit our evaluation to supervised OIE systems.

We evaluate our system's performance against the gold test data in \textsc{lsoie}-wiki and \textsc{lsoie}-sci by considering extractions to be a match if they contain the same predicate as the gold extraction and contain the syntactic head of each gold argument. Syntactic heads are extracted with the Stanford CoreNLP dependency parser \cite{chen-manning-2014-fast}. Although it would be ideal to have the gold syntactic head, this method is preferable to taking the lexical overlap of the entire extraction \citet{stanovsky-dagan-2016-creating}, ignoring argument tags and ordering as pointed out in \citet{lechelle-etal-2019-wire57}.  

We then assign a confidence score to each extraction to allow for tuning the precision-recall tradeoff. For the non-CRF models, we use the mean log probability assigned to the tag labels in the extraction as the confidence score. For the CRF model, we use the log probability assigned to the entire sequence. We differ from \citet{stanovsky-etal-2018-supervised} where confidence was calculated as the product of the inverse of the model's estimate probability for each tag label, preferring longer extractions which were more likely to get a 50\% lexical match, outweighing the deficit of swimming upstream against the model's estimated confidence and still producing a downward sloping precision recall curve. 

We use Viterbi decoding to extract the most likely valid BIO tagging sequence given the model's probability output for each BIO tag. We import the Viterbi algorithm functionality from the AllenNLP library \citep{DBLP:journals/corr/abs-1803-07640}.

\section{Discussion}
Figure~\ref{fig:results-graph} shows precision and recall curves on the \textsc{lsoie}-wiki test set, accompanied by the \texttt{ls\_oie} model's estimated confidence. Table~\ref{tab:results-table} shows $F_1$ and AUC scores for the benchmark models on the \textsc{lsoie}-wiki and \textsc{lsoie}-sci test sets. 

The OIE modeling task is difficult. Results on both evaluation sets show that
the BERT model and the CRF output layer
improve over the baseline model. 
Training with the \textsc{lsoie} improves model performance.
When science is the target domain, the \texttt{*\_sci} models are preferable, as they have slightly higher in-domain performance, showing the value of the domain split in \textsc{lsoie}.

\subsection{Error Analysis}
We conduct a manual error analysis of the \texttt{ls\_oie} model, where we find that our baseline models could benefit from more careful extractions.

\textbf{Incorrect predicate:}
At minimum confidence, 53\% of the model's precision errors come from verbs that are not present in the gold dataset.
Half of these are legitimate predicates that are missing from the gold dataset and the other half are auxiliary verbs, that should not be present in the gold dataset. Depending on the deployment environment, the model could be improved with predicate filtering heuristics at prediction time.

\textbf{Argument Concatenation:}
We examined 500 incorrect extractions by \texttt{ls\_oie}. We found that 36\% of unmatched extractions
were semantically similar to the gold extraction.
These extractions either concatenated arguments $A_1$-$A_N$ into $A_1$ while gold did not, split these arguments apart while gold did, or dropped a non-material argument. For future modeling, this is an argument to drop $A_2$ and beyond from the dataset and only model OIE with extraction triples.

\textbf{True Errors:}
Among the extraction errors, 2/3 involve errors in argument ordering, often following the natural order of the sentence.
The other 1/3 of errors involved the model making nonsensical extractions or not making extracting arguments beyond $A_0$, presumably because of lack of confidence and defaulting to the $O$ label.

\textbf{LSOIE Modeling Improvements:}
We also manually examined 100 extractions where \texttt{ls\_oie} chose the right extraction over \texttt{rnnoie}. In these cases, we found improved argument ordering, increased confidence on relevant $A_1$ objects, and better accuracy identifying subjects that are distant from the predicate.

\section{Conclusion}

In this paper, we introduced the \textsc{lsoie} dataset as a resource for supervised OIE.
We have algorithmically re-purposed the \textsc{qa-srl bank} 2.0 into a new OIE dataset, \textsc{lsoie}, which contains over 70,000 sentences and over 150,000 extraction tuples.
To benchmark the new dataset, we trained and evaluated a series of supervised OIE models, providing baselines for future research on the OIE modeling task.

The code and datasets introduced in this paper can be found at \url{https://github.com/Jacobsolawetz/large-scale-oie}.

\section*{Acknowledgments}
A special thanks to Julian Michael and Gabriel Stanovsky for providing much needed guidance and feedback for our research. This project would not have been possible without their insight and previous progress on the topic. We also thank the anonymous reviewers for their feedback.


\bibliographystyle{acl_natbib}

\end{document}